\documentclass{article}

\usepackage[final]{nips_2018}

\usepackage[utf8]{inputenc} 
\usepackage[T1]{fontenc}    
\usepackage{hyperref}       
\usepackage{url}            
\usepackage{booktabs}       
\usepackage{amsfonts}       
\usepackage{nicefrac}       
\usepackage{microtype}      
\usepackage{natbib}
\usepackage{url}
\usepackage{breakurl}

\title{Law and Adversarial Machine Learning}

\author{
  Ram Shankar Siva Kumar\\
Microsoft\\
  \texttt{Ram.Shankar@microsoft.com} \\
  \And
 David R. O'Brien \\
  Berkman Klein Center for Internet and Society\\
  \texttt{dobrien@cyber.harvard.edu} \\
  \And
  Kendra Albert \\
  Harvard Law School\\
  \texttt{kalbert@law.harvard.edu} \\
  \And
  Salom\'e Viljoen  \\
  Berkman Klein Center for Internet and Society\\
  \texttt{sviljoen@cyber.harvard.edu} \\
}

\begin{document}

\maketitle

\begin{abstract}
When machine learning systems fail because of adversarial manipulation, how should society expect the law to respond? Through scenarios grounded in adversarial ML literature, we explore how some aspects of computer crime, copyright, and tort law interface with perturbation, poisoning, model stealing and model inversion attacks to show how some attacks are more likely to result in liability than others. We end with a call for action to ML researchers to invest in transparent benchmarks of attacks and defenses; architect ML systems with forensics in mind and finally, think more about adversarial machine learning in the context of civil liberties. The paper is targeted towards ML researchers who have no legal background.
\end{abstract}

\section{Introduction}
Technology and the law are inextricably linked, and for either to be effective for society, the two must work together.  As new technologies permit new potential harms, judges, legislatures, and regulators are on the spot for rationalizing law with technology. For adversarial machine learning, this process is just beginning - judges have not had much cause to determine the applicability of existing law to attacks on machine learning, nor have specific laws been passed to regulate machine learning systems.

But the arms-length relationship between Machine Learning (ML) attacks and the law seem unlikely to continue. Despite the vulnerabilities that this community has demonstrated since 2004, as \citet{biggio2018wild} notes, ML is at the core of many critical systems including healthcare, defense, and finance . Given this, when such systems fail or get compromised, it seems inevitable that law and adversarial machine learning are on a crash course towards each other. But the relationship is under-theorized. In response to \citet{tramer2016stealing} work on model stealing, one of the affected companies responded: “Said another way, even if stealing software were easy, there is still an important disincentive to do so in that it violates intellectual property law.” (see  \citet{cetinsoy_2016}) Such a statement assumes, without proof, that model stealing can be sanctioned by existing intellectual property law. As we discuss below, it is entirety possible that it won't be. \textit{The goal of this paper is to begin to explore how for some attacks, existing law may provide protection for ML models, but in others, there may be less protection for machine learning systems than practitioners expect.}

 We have structured the paper thus: we discuss supply chain, perturbation and poisoning attacks via the Computer Fraud and Abuse Act (Section II); model stealing and model inverstion in the lens of U.S. intellectual property law (Section III); applicability of civil liability law to adversarial ML (Section IV), finally ending with some recommendations for ML researcher (Section V). Since our paper is tailored to the ML community who have no legal background, we only provide a sampling of the legal concepts as it relates to ML attacks.

\section{Cybersecurity law and Supply Chain, Perturbation, Poisoning attacks}
In this section we discuss supply chain, perturbation and poisoning attacks through the lens of the Computer Fraud and Abuse Act (CFAA), the hallmark federal “anti-hacking” statute in the US. CFAA was originally enacted in 1984, and inspired in part by the film "War Games" and has surprisingly kept up with 30 years of technological changes. Simply stated, the CFAA broadly prohibits individuals from intentionally accessing computers without authorization, exceeding authorized access on a computer, and causing damage to computers without authorization.  Violators of these provisions may face lawsuits by the victims and criminal prosecution. US attorneys have successfully used the law to prosecute a wide range of activities – some would argue too expansively – thanks in large part to its broadly-worded prohibitions as noted by \citet{curtiss2016computer}. That said, adversarial ML might be different.  As \citet{calo2018tricking} point out, adversarial ML attacks present definitional challenges that raise questions about whether the CFAA is up to task. Much of the CFAA is couched around whether access has occurred or a system damaged.The scenarios we explore below are intended to selectively illustrate both parity and disparity that might arise between the CFAA and an attack.

\underline{Scenario:} \citet{gu2017badnets} propose attackers may target the ML supply chain by compromising the pre-trained models as they are downloaded from an insecure (HTTP) connection.\\
\underline{Legal commentary:} A classic man-in-the-middle attack like this appears to be a straight-forward CFAA violation -- the attacker knowingly accessed and altered the model in transit without authorization. Similarly, an attacker exploiting a buffer overflow vulnerability on OpenCV that results in misclassification as demonstrated by \citet{xiao2017security}, likely violates the CFAA, since the attacker has accessed the platform and altered the integrity of the output by exploitation. 

\underline{Scenario:} \citet{jagielski2018manipulating} poison a healthcare dataset quite effectively that a tenth of the patients have their dosages changed by 359\%. \\ 
\underline{Legal commentary:} A poisoning attack like this could plausibly be a violation of the CFAA.  The strongest argument may be that in carrying out the attack, the adversary transmitted a code (in this context, prosecutors could argue that code is the poisoned examples) that caused damage to the model in a way that disrupts the system.  However, the analysis becomes less clear in cases where the purpose of the ML system is more open-ended or premised on interactive feedback, like Tay. When in principle do innocuous inputs become malicious? And, at what point does an ML system reach a state of being damaged?    

\underline{Scenario:} \citet{papernot2015distillation} propose to fool a bank’s image recognition system to misrecognize checks to higher value. \\ 
\underline{Legal commentary:} Perturbation attacks may be covered under the CFAA as prosecutors could argue that the adversary knowingly transmitted code (in this case a modified image, which ultimately gets converted to code) that caused damage (in this case monetary damages suffered by the bank). By the same token, it can also be argued that tampering with stop signs in the context of autonomous cars is also a CFAA violation, since stickers (as seen in  \citet{evtimov2017robust} ) are a form of transmitting code that causes damage to the autonomous car (a computer in the eyes of CFAA). \\
Banks (and other organizations) are also likely to have a Terms of Services (ToS) drafted by its lawyers which generally prohibit malicious activities. Several CFAA cases have turned on whether the activities in question were prohibited by a ToS agreement, which some courts have held can constitute exceeding authorized access under the CFAA.  However, it is a controversial subject.  In situations where the applicability of the CFAA may not be clear based on the statutory definitions, a ToS can bridge some of these gaps. Finally, since it is common for prosecutors to pursue higher charges, in addition to CFAA violation the adversary would also face wire fraud charges.  \\

The takeaway from this section should be that the CFAA plausibly covers some supply chain, perturbation and poisoning attacks if its statutory language is interpreted in certain ways. Courts have struggled in similar cases in the past, as \citet{calo2018tricking} discuss in more detail, and it is far from clear whether they can consistently resolve these differences in future cases.

\section{Copyright law and Model Inversion, Model Extraction attacks}
To protect against model inversion and model stealing, ML practitioners may be tempted to turn to a different body of law – copyright law. However, unlike CFAA which is broad and open to wide interpretation,  copyright law is more narrow and well-defined, and hence unlikely to provide as much coverage as the CFAA.

\underline{Scenario:} \citet{fredrikson2015model} reconstruct part of the private training data using hill climbing on output probabilities  \\
\underline{Legal Commentary:} The ability of the owner, whose data was reconstructed, to get relief to under copyright would depend upon what exactly the training data was. In the United States, facts are not copyrightable, even if they are costly or time-consuming to gather. Copyright protection may attach to compilations or arrangements of factual information, however, it is unlikely that a reconstructed set of training data would necessarily share the same compilation or arrangement as the original: for instance, the reconstructed data could be approximations as in the case of \citet{fredrikson2015model}. So the owner of the dataset/model would be unlikely to be able to successfully sue an adversary that recovered part of a training dataset consisting of facts for copyright infringement. \\
On the other hand, images and audio are copyrightable, so, the owner would be more likely to succeed against an adversary that reproduced those. The question is murkier with regards to information derived from copyrightable materials, such as RGB pixel values, or general image characteristics. 

\underline{Scenario:} \citet{tramer2016stealing} reconstruct a model hosted behind a prediction API  \\
\underline{Legal Commentary:} Copyright for software is an interest in a code as a “literary work”, not for its functions. Therefore, although the code that runs a particular machine learning model might be protected by copyright, a reconstruction is unlikely to share the particular expression of code with the original, and thus reconstruction is unlikely to violate copyright law: for instance, even if both the original and the “stolen” model are decision trees as shown in \citet{tramer2016stealing}, their exact implementation may differ and thus the “stolen” model would not infringe copyright. \\
It is possible that in some circumstances, machine learning models may qualify as trade secrets, and that trade secret law could protect a model against reconstruction. However, in order to successfully sue for trade secret disclosure, an owner must show that they took reasonable precautions to prevent disclosure.

In the absence of intellectual property protection, one potential way to prevent model stealing would be to include a Terms of Service that specifically prohibits this, thereby establishing a contract with the API users. However, such contractual agreements only create rights against the users of the API – they might not help if an adversary releases a reconstructed model publicly. But in any case, model inversion and model extraction are attacks where existing law might not protect ML systems in the way that companies or researchers might expect.

\section{Liability laws in the context of adversarial ML}
In this section we attempt to answer the following question using tort law: When an ML product breaks down because of adversarial examples, who is liable? This question is not purely rhetorical: the European Union is set to release a liability and safety framework for ML systems by mid-2019 (see \citet{digital_single_market_2018})  which could snowball into GDPR style regulation.

\underline{Scenario:} \citet{brundage2018malicious} discuss how a drone’s image recognition system could fail owing to adversarial examples and potentially cause damage. While the authors discuss this in the context of military drones, we will assume that the drone is consumer grade (as used in photography) to avoid complications with international laws. \\
\underline{Legal Commentary:} The uncertainty here arises due to the interaction between a vulnerable product and a malicious actor. \citet{gilmer2018motivating} argue that as long as there is non-zero test error, adversarial examples will exist. If a drone vendor used a state of the art image recognition system (which is likely to have non-zero test error), was the manufacturer negligent? Software vendors have generally not been held liable for traditional software attacks under theories of product liability. Yet the novel nature and expanded scope of harms presented by ML products may pose new risks for this type of liability.

Part of the issue is that courts do not have industry standards with which to compare negligent versus responsible ML development practices. No established standard or industry wide practice for protecting against adversarial examples or reward hacking has been established. Another complicating factor is the interrelated nature of the ML ecosystem makes it difficult to establish which component caused the failure. Machine learning systems are built on a mix of open source libraries and commercial systems. For example, consider the (common) case wherein a vendor, say ,a drone manufacturer, reuses a model from academic researchers hosted in Caffee Model Zoo, ports it over in PyTorch and runs it on commercial cloud. When the drone fails and causes bodily harm, who is liable? The answer, as noted by \citet{calo2010open} is not known and we may have to wait until such a case comes to trial to provide some insight into how blame will be assigned in this ecosystem.

\section{Call to Action for ML Researchers: }
Given the uncertainty in law in some adversarial ML attacks, here are three recommendations for ML researchers working in this space to assist lawyers and policy makers in creating reasonable, evidence-driven policy:
\begin{enumerate}
\item Benchmark Attacks and Defenses –  There is a growing need for legal practioners and policy makers to understand how adversarial ML differs from traditional software attacks in ways that may inform how laws are interpreted and enforced.  ML researchers can bring clarity to the situation by helping to prioritize the attacks and defenses they publish. This will both help inform the development of appropriate standards of care for systems that use machine learning, and provide practical guidance to engineers.  
\subitem In this spirit, whenever researchers publish a new defense against an attack, they might consider using tools like cleverhans (See \citet{ papernot2018cleverhans}) , IBM's adversarial robustness toolkit \citet{art2018} and report shortcomings. The community should expand and invest in benchmarking efforts such as RobustML (see \citet{robust2009}). We found \cite{goodfellow2018defense}, where defenses are stack ranked, useful to think about the progress of defenses in perturbation attacks. 
\subitem Benchmarks alone aren't sufficient: we also think there is a need for a framework to assess risk and prioritize adversarial ML threats realistically. Attackers need not perform perturbation attacks to evade ML systems as documented by \citet{gilmer2018motivating}. To address this we believe the ML community can take inspiration from threat modeling from software community \textbf{DREAD} (see \citet{Shostack2014}) to prioritize software threats. It rates attacks based on the potential for \textbf{D}amage, \textbf{R}eliability of attack, the ease which an attacker can launch the \textbf{E}xploit, the scope of \textbf{A}ffected users, and the ease with which an attacker can \textbf{D}iscover the attack. 
\item Architect for forensics – ML systems are currently not built with forensics in mind. From a legal perspective, forensics can lend clues to attack attribution and hence eventual prosecution. ML researchers should be thinking proactively about how to architect systems so that investigations are possible, including mechanisms to alert when the system is under adversarial attack, recommend appropriate logging, construct playbooks for incident response during an attack and formulate remediation plan to recover to from the adversarial attack. 
\item Take into account civil liberties - Deployed ML systems have the ability to impact civil liberties and basic human rights such as freedom of expression and privacy. For instance, ML researchers should anticipate that oppressive governments could seek backdoors in consumer ML systems, as demonstrated by \citet{chen2017}, facilitate censorship and out political dissidents. On the same note, researchers must also think about the \emph{dual use of adversarial examples} i.e. the benefits of adversarial examples. For example, dissidents in a totalitarian state should be able to evade facial detection using 3D printed glasses as shown by \citet{sharif2017adversarial}. ML researchers should do their best to anticipate how ML systems and attacks can be used for the benefit and detriment of individuals. 
\end{enumerate}

\section{Conclusion}
Given the widespread usage of ML in real world applications, legal responses to adversarial ML attacks are important to society and inevitable. Some aspects of the law map onto attacks such as poisoning and perturbation, but for others, like model stealing, legal recourse is less clear. ML practitioners can bring clarity to this discussion by consindering benchmarking attacks and defenses; architecting ML systems with in built forensics, and be thoughtful about the \emph{dual use of adversarial examples}

\subsubsection*{Acknowledgments}

 An interdisciplinary paper such as this would not have been possible without fruitful discussions and feedback from ML researchers (Aleksandr Madry, Momin Malik, Gretchen Greene, Sharon Gillett, Justin Gilmer), security experts (John Walton, John Lambert, Jeffrey Snover, Matt Swann) and  lawyers/public policy experts (Woodrow Hartzog, Daniel Edelman, Ryan Calo, Yaniv Benhamou, Cristin Goodwin). Ram would also like to thank Andi Comissoneru, Sharon Xia, Steve Mott and the entire Azure Security Data Science team for holding the fort during his time away.  

\medskip

\small

\bibliographystyle{plainnat}
\bibliography{refLawAndAdvML}

\end{document}